\documentstyle[eaclap]{article}
\newcommand 	{\cky}		{{\small \sf CKY}}
\newcommand 	{\cfg}		{{\small \sf CFG}}
\newcommand 	{\lig}		{{\small \sf LIG}}
\newcommand 	{\ig}		{{\small \sf IG}}

\newcommand 	{\plig}		{{\small \sf PLIG}}
\newcommand 	{\pltg}		{{\small \sf PLTG}}
\newcommand 	{\PATR}		{{\small \sf PATR}}
\newcommand 	{\plpatr}	{{\small \sf PLPATR}}

\newcommand	{\stck}		{\insqr}

\newcommand	{\tree}		{\insqr}

\def        	\buildrell#1\over#2{\mathrel{\mathop{\null#1}\limits_{#2}}}
\newcommand	{\derives}[2]	{\buildrel{#2}\over{\buildrell\Longrightarrow
					\over{\hbox{\raise5pt
					\hbox{$_{_{#1}}$}}}}}

\newcommand	{\set}[1]	{\left\{\,{#1}\,\right\}}

\newcommand	{\setof}[2]	{\left\{\,#1\,\left|\,
					\mbox{#2}\right.\,\right\}}
\newcommand	{\insqr}[1]	{{\left[{#1}\right]}}

\newcommand{\comment}[1]{\mbox{}}

\newtheorem{Exam}{Example}%[section]
\newenvironment{Example}{\Exam}{\endExam\smallskip}
\newcommand{\boxed}[1] {\fbox{#1}}
\newcommand{\fs}[1]
  {{\tiny\left[\begin{array}{l}#1\end{array}\right]}}

\newcommand{\rfs}[2]
  {{\tiny\left[\begin{array}{l}#1\end{array}\right]^{\boxed{#2}}}}

\newcommand{\fsnorm}[1]
  {\left[\begin{array}{l}#1\end{array}\right]}

\newcommand{\rootfs}
{\fs{f:\rfs{f:a\\g:F_1}{1}\\
     g:\rfs{f:F_2\\g:b}{2}}}

\newcommand{\rootfsb}
{\fs{f:\rfs{f:a\\g:F'_1}{1}\\
     g:\rfs{f:F'_2\\g:b}{2}}}

\newcommand{\ftermfs}
{\rfs{f:a\\
      g:\fs{f:c\\g:F_3}}{1}}

\newcommand{\gtermfs}
{\rfs{f:\fs{f:F_4\\g:d}\\
      g:b}{2}}

\newcommand{\ftermfsa}
{\fs{f:a\\
      g:\fs{f:c'\\g:F'_3}}}

\newcommand{\gtermfsa}
{\fs{f:\fs{f:F'_4\\g:d'}\\
      g:b}}

\newcommand{\ftermfsb}
{\rfs{f:a\\
      g:\fs{f:c'\\g:F'_3}}{1}}

\newcommand{\gtermfsb}
{\rfs{f:\fs{f:F'_4\\g:d'}\\
      g:b}{2}}

\newcommand{\hidden}[1]{\mbox{}}
\newcommand{\mlines}[1]{\begin{tabular}[c]{c}#1\end{tabular}}

\newcommand{\leftmostsym}{\mlines{first\\nonblank\\symbol}}
\newcommand{\rightmostsym}{\mlines{last\\nonblank\\symbol}}

\author{Bill Keller \and David Weir \\ School
of Cognitive and Computing Sciences \\ University of Sussex \\
Falmer, Brighton BN1 9QH\\ UK\\
{\tt bill.keller/david.weir@cogs.sussex.ac.uk}}
\title{A Tractable Extension of Linear Indexed Grammars}
\begin{document}

\maketitle
\bibliographystyle{acl}

\begin{abstract}

Vijay-Shanker and Weir \shortcite{vw93b} show that Linear Indexed
Grammars (\lig) can be processed in polynomial time by exploiting
constraints which make possible the extensive use of
structure-sharing. This paper describes a formalism that is more
powerful than \lig, but which can also be processed in polynomial time
using similar techniques. The formalism, which we refer to as Partially
Linear \PATR\ (\plpatr) manipulates feature structures rather than
stacks.

\end{abstract}

\section{Introduction}

Unification-based grammar formalisms can be viewed as generalizations of
Context-Free Grammars~(CFG) where the nonterminal symbols are replaced
by an infinite domain of feature structures.  Much of their popularity
stems from the way in which syntactic generalization may be elegantly
stated by means of constraints amongst features and their values.
Unfortunately, the expressivity of these formalisms can have undesirable
consequences for their processing.  In naive implementations of
unification grammar parsers, feature structures play the same role as
nonterminals in standard context-free grammar parsers. Potentially large
feature structures are stored at intermediate steps in the computation,
so that the space requirements of the algorithm are
expensive. Furthermore, the need to perform non-destructive unification
means that a large proportion of the processing time is spent copying
feature structures.

One approach to this problem is to refine parsing algorithms by
developing techniques such as restrictions, structure-sharing, and lazy
unification that reduce the amount of structure that is stored and hence
the need for copying of features
structures~\cite{sh85c,p85,kk85b,wrob87,gerd89,god90,kog90,eme91,toma91,harrell92}). While
these techniques can yield significant improvements in performance, the
generality of unification-based grammar formalisms means that there are
still cases where expensive processing is unavoidable. This approach
does not address the fundamental issue of the tradeoff between the
descriptive capacity of a formalism and its computational power.

In this paper we identify a set of constraints that can be placed on
unification-based grammar formalisms in order to guarantee the existence
of polynomial time parsing algorithms.  Our choice of constraints is
motivated by showing how they generalize constraints inherent in Linear
Indexed Grammar (\lig). We begin by describing how constraints inherent
in \lig\ admit tractable processing algorithms and then consider how
these constraints can be generalized to a formalism that manipulates
trees rather than stacks. The constraints that we identify for the
tree-based system can be regarded equally well as constraints on
unification-based grammar formalisms such as \PATR\ \cite{sh84}.

\section{From Stacks to Trees}

An Indexed Grammar ({\ig}) can be viewed as a {\cfg} in which each
nonterminal is associated with a stack of indices.  Productions specify
not only how nonterminals can be rewritten but also how their associated
stacks are modified.  {\lig}, which were first described by
Gazdar~\shortcite{gaz88}, are constrained such that stacks are passed
from the mother to at most a single daughter.

For \lig, the size of the domain of nonterminals and associated stacks
(the analogue of the nonterminals in \cfg) is not bound by the grammar.
However, Vijay-Shanker and Weir~\shortcite{vw93b} demonstrate that
polynomial time performance can be achieved through the use of
structure-sharing made possible by constraints in the way that \lig\ use
stacks.  Although stacks of unbounded size can arise during a
derivation, it is not possible for a {\lig} to specify that two
dependent, unbounded stacks must appear at distinct places in the
derivation tree. Structure-sharing can therefore be used effectively
because checking the applicability of rules at each step in the
derivation involves the comparison of structures of limited size.

Our goal is to generalize the constraints inherent in \lig\, to a
formalism that manipulates feature structures rather than stacks.  As a
guiding heuristic we will avoid formalisms that generate tree sets with
an {\em unbounded\/} number of unbounded, dependent branches.
It appears that the structure-sharing techniques used with \lig\ cannot
be generalized in a straightforward way to such formalisms.

Suppose that we generalize \lig\ to allow the stack to be passed from
the mother to two daughters. If this is done recursion can be used to
produce an unbounded number of unbounded, dependent branches.  An
alternative is to allow an unbounded stack to be shared between two (or
more) daughters but {\em not\/} with the mother.  Thus, rules may
mention more than one unbounded stack, but the stack associated with the
mother is still associated with at most one daughter. We refer to this
extension as Partially Linear Indexed Grammars~({\plig}).

\begin{Example}
The \plig\ with the following productions
generates the language $$\setof{a^nb^mc^nd^m}{$n,m\ge 1$}$$ and the tree
set shown in Figure~\ref{fig-pliga}.  Because a single \plig\ production
may mention more than one unbounded stack, variables ($x$, $y$) are
introduced to distinguish between them. The notation $A\stck{x\sigma}$
is used to denote the nonterminal $A$ associated with any stack whose
top symbol is $\sigma$.

\[
\begin{array}[c]{ll} A\stck{x}\rightarrow
aA\stck{x\sigma},\qquad & A\stck{x}\rightarrow
B\stck{y}C\stck{x}D\stck{y},\\ B\stck{x\sigma}\rightarrow bB\stck{x},
& B\stck{\sigma}\rightarrow b,\\ C\stck{x\sigma}\rightarrow
cC\stck{x}, & C\stck{\sigma}\rightarrow c,\\
D\stck{x\sigma}\rightarrow dD\stck{x},  &
D\stck{\sigma}\rightarrow d.
\end{array}
\]
\end{Example}

\begin{figure}
\begin{center}
\setlength{\unitlength}{0.012500in}%
\begingroup\makeatletter\ifx\SetFigFont\undefined
% extract first six characters in \fmtname
\def\x#1#2#3#4#5#6#7\relax{\def\x{#1#2#3#4#5#6}}%
\expandafter\x\fmtname xxxxxx\relax \def\y{splain}%
\ifx\x\y   % LaTeX or SliTeX?
\gdef\SetFigFont#1#2#3{%
  \ifnum #1<17\tiny\else \ifnum #1<20\small\else
  \ifnum #1<24\normalsize\else \ifnum #1<29\large\else
  \ifnum #1<34\Large\else \ifnum #1<41\LARGE\else
     \huge\fi\fi\fi\fi\fi\fi
  \csname #3\endcsname}%
\else
\gdef\SetFigFont#1#2#3{\begingroup
  \count@#1\relax \ifnum 25<\count@\count@25\fi
  \def\x{\endgroup\@setsize\SetFigFont{#2pt}}%
  \expandafter\x
    \csname \romannumeral\the\count@ pt\expandafter\endcsname
    \csname @\romannumeral\the\count@ pt\endcsname
  \csname #3\endcsname}%
\fi
\fi\endgroup
\begin{picture}(265,293)(427,485)
\thinlines
\put(560,760){\line( 0,-1){ 20}}
\put(560,760){\line(-2,-1){ 40}}
\put(560,720){\line(-2,-1){ 40}}
\multiput(560,720)(0.00000,-8.00000){3}{\line( 0,-1){  4.000}}
\put(560,680){\line( 0,-1){ 20}}
\put(560,680){\line(-2,-1){ 40}}
\put(560,640){\line( 0,-1){ 20}}
\put(600,645){\makebox(0.1111,0.7778){\SetFigFont{5}{6}{rm}.}}
\put(480,600){\line( 0,-1){ 20}}
\put(480,560){\line(-2,-1){ 40}}
\put(480,600){\line(-2,-1){ 40}}
\put(560,640){\line(-4,-1){ 80}}
\put(560,640){\line( 4,-1){ 80}}
\put(560,600){\line( 0,-1){ 20}}
\put(560,600){\line(-2,-1){ 40}}
\put(560,560){\line(-2,-1){ 40}}
\put(640,600){\line(-2,-1){ 40}}
\put(640,600){\line( 0,-1){ 20}}
\multiput(480,560)(0.00000,-8.00000){3}{\line( 0,-1){  4.000}}
\multiput(560,560)(0.00000,-8.00000){3}{\line( 0,-1){  4.000}}
\multiput(640,560)(0.00000,-8.00000){3}{\line( 0,-1){  4.000}}
\put(475,520){\line(-2,-1){ 40}}
\put(555,520){\line(-2,-1){ 40}}
\put(640,560){\line(-2,-1){ 40}}
\put(640,520){\line(-2,-1){ 40}}
\put(560,725){\makebox(0,0)[b]{\smash{\SetFigFont{10}{12.0}{rm}$A\stck{\sigma}$}}}
\put(520,725){\makebox(0,0)[b]{\smash{\SetFigFont{10}{12.0}{rm}$a$}}}
\put(520,685){\makebox(0,0)[b]{\smash{\SetFigFont{10}{12.0}{rm}$a$}}}
\put(560,765){\makebox(0,0)[b]{\smash{\SetFigFont{10}{12.0}{rm}$A\stck{}$}}}
\put(560,685){\makebox(0,0)[b]{\smash{\SetFigFont{10}{12.0}{rm}$A\stck{\sigma^{n-1}}$}}}
\put(560,645){\makebox(0,0)[b]{\smash{\SetFigFont{10}{12.0}{rm}$A\stck{\sigma^n}$}}}
\put(520,645){\makebox(0,0)[b]{\smash{\SetFigFont{10}{12.0}{rm}$a$}}}
\put(480,605){\makebox(0,0)[b]{\smash{\SetFigFont{10}{12.0}{rm}$B\stck{\sigma^m}$}}}
\put(480,565){\makebox(0,0)[b]{\smash{\SetFigFont{10}{12.0}{rm}$B\stck{\sigma^{m-1}}$}}}
\put(440,525){\makebox(0,0)[b]{\smash{\SetFigFont{10}{12.0}{rm}$b$}}}
\put(440,565){\makebox(0,0)[b]{\smash{\SetFigFont{10}{12.0}{rm}$b$}}}
\put(560,605){\makebox(0,0)[b]{\smash{\SetFigFont{10}{12.0}{rm}$C\stck{\sigma^n}$}}}
\put(520,525){\makebox(0,0)[b]{\smash{\SetFigFont{10}{12.0}{rm}$c$}}}
\put(560,565){\makebox(0,0)[b]{\smash{\SetFigFont{10}{12.0}{rm}$C\stck{\sigma^{n-1}}$}}}
\put(520,565){\makebox(0,0)[b]{\smash{\SetFigFont{10}{12.0}{rm}$c$}}}
\put(640,605){\makebox(0,0)[b]{\smash{\SetFigFont{10}{12.0}{rm}$D\stck{\sigma^m}$}}}
\put(640,565){\makebox(0,0)[b]{\smash{\SetFigFont{10}{12.0}{rm}$D\stck{\sigma^{m-1}}$}}}
\put(600,525){\makebox(0,0)[b]{\smash{\SetFigFont{10}{12.0}{rm}$d$}}}
\put(600,565){\makebox(0,0)[b]{\smash{\SetFigFont{10}{12.0}{rm}$d$}}}
\put(475,525){\makebox(0,0)[b]{\smash{\SetFigFont{10}{12.0}{rm}$B\stck{\sigma}$}}}
\put(435,485){\makebox(0,0)[b]{\smash{\SetFigFont{10}{12.0}{rm}$b$}}}
\put(515,485){\makebox(0,0)[b]{\smash{\SetFigFont{10}{12.0}{rm}$c$}}}
\put(555,525){\makebox(0,0)[b]{\smash{\SetFigFont{10}{12.0}{rm}$C\stck{\sigma}$}}}
\put(635,525){\makebox(0,0)[b]{\smash{\SetFigFont{10}{12.0}{rm}$D\stck{\sigma}$}}}
\put(595,485){\makebox(0,0)[b]{\smash{\SetFigFont{10}{12.0}{rm}$d$}}}
\end{picture}
\end{center}
\caption{Tree set for  $\setof{a^nb^mc^nd^m}{$n,m\ge 1$}$}
\label{fig-pliga}
\end{figure}

\begin{Example}
A \plig\ with the following productions generates the $k$-copy
language over $\set{a,b}^*$, i.e., the language
\[\setof{w^k}{$w\in\set{a,b}^*$}\]
where $k\ge 1$.
\[
\begin{array}[c]{ll} S\stck{}\rightarrow
\underbrace{A\stck{x}\ldots A\stck{x}}_{\mbox{$k$ copies}}, &
A\stck{}\rightarrow\lambda,\\\\
A\stck{x\sigma_1}\rightarrow a\,A\stck{x}, &
A\stck{x\sigma_2}\rightarrow b\,A\stck{x}.
\end{array}
\]
\label{ex-kcopy}
\end{Example}

\begin{Example}
\plig\ can ``count'' to any fixed $k$, i.e., a \plig\ with the following
productions generates the language
\[\setof{a_1^n\ldots a_k^n}{$n\ge 0$}\] where $k\ge 1$.
\[
\begin{array}[c]{ll} S\stck{}\rightarrow
A_1\stck{x}\ldots A_k\stck{x},\\
A_1\stck{x\sigma}\rightarrow a_1\,A_1\stck{x},&
A_1\stck{}\rightarrow\lambda, \\
\vdots\\
A_k\stck{x\sigma}\rightarrow a_k\,A_k\stck{x},&
A_k\stck{}\rightarrow\lambda.\\
\end{array}
\]
\label{ex-kcount}
\end{Example}

In \plig, stacks shared amongst siblings cannot be passed to the
mother. As a consequence, there is no possibility that recursion can be
used to increase the number of dependent branches. In fact, the number
of dependent branches is bounded by the length of the right-hand-side of
productions.  By the same token, however, \plig\ may only generate
structural descriptions in which dependent branches begin at nodes that
are siblings of one another. Note that the tree shown in
Figure~\ref{fig-pltg} is unobtainable because the branch rooted at
$\eta_1$ is dependent on more than one of the branches originating at
its sibling $\eta_2$.

\begin{figure}
\begin{center}
\setlength{\unitlength}{0.012500in}%
\begingroup\makeatletter\ifx\SetFigFont\undefined
% extract first six characters in \fmtname
\def\x#1#2#3#4#5#6#7\relax{\def\x{#1#2#3#4#5#6}}%
\expandafter\x\fmtname xxxxxx\relax \def\y{splain}%
\ifx\x\y   % LaTeX or SliTeX?
\gdef\SetFigFont#1#2#3{%
  \ifnum #1<17\tiny\else \ifnum #1<20\small\else
  \ifnum #1<24\normalsize\else \ifnum #1<29\large\else
  \ifnum #1<34\Large\else \ifnum #1<41\LARGE\else
     \huge\fi\fi\fi\fi\fi\fi
  \csname #3\endcsname}%
\else
\gdef\SetFigFont#1#2#3{\begingroup
  \count@#1\relax \ifnum 25<\count@\count@25\fi
  \def\x{\endgroup\@setsize\SetFigFont{#2pt}}%
  \expandafter\x
    \csname \romannumeral\the\count@ pt\expandafter\endcsname
    \csname @\romannumeral\the\count@ pt\endcsname
  \csname #3\endcsname}%
\fi
\fi\endgroup
\begin{picture}(252,293)(214,525)
\put(240,705){\makebox(0,0)[b]{\smash{\SetFigFont{10}{12.0}{rm}$a$}}}
\put(240,665){\makebox(0,0)[b]{\smash{\SetFigFont{10}{12.0}{rm}$a$}}}
\put(240,625){\makebox(0,0)[b]{\smash{\SetFigFont{10}{12.0}{rm}$a$}}}
\put(380,585){\makebox(0,0)[b]{\smash{\SetFigFont{10}{12.0}{rm}$c$}}}
\put(380,625){\makebox(0,0)[b]{\smash{\SetFigFont{10}{12.0}{rm}$c$}}}
\put(380,545){\makebox(0,0)[b]{\smash{\SetFigFont{10}{12.0}{rm}$c$}}}
\put(310,665){\makebox(0,0)[b]{\smash{\SetFigFont{10}{12.0}{rm}$b$}}}
\put(310,625){\makebox(0,0)[b]{\smash{\SetFigFont{10}{12.0}{rm}$b$}}}
\put(310,585){\makebox(0,0)[b]{\smash{\SetFigFont{10}{12.0}{rm}$b$}}}
\thinlines
\put(310,780){\line(-2,-1){ 40}}
\put(270,740){\line( 0,-1){ 20}}
\put(410,660){\line( 0,-1){ 20}}
\put(310,780){\line( 4,-1){ 80}}
\put(390,740){\line(-5,-2){ 50}}
\put(390,800){\vector( 0,-1){ 30}}
\put(270,800){\vector( 0,-1){ 30}}
\put(340,700){\line( 0,-1){ 20}}
\multiput(270,700)(0.00000,-8.00000){3}{\line( 0,-1){  4.000}}
\multiput(340,660)(0.00000,-8.00000){3}{\line( 0,-1){  4.000}}
\multiput(410,620)(0.00000,-8.00000){3}{\line( 0,-1){  4.000}}
\put(270,740){\line(-3,-2){ 30}}
\put(270,700){\line(-3,-2){ 30}}
\put(270,660){\line(-3,-2){ 30}}
\put(340,700){\line(-3,-2){ 30}}
\put(340,660){\line(-3,-2){ 30}}
\put(340,620){\line(-3,-2){ 30}}
\put(410,660){\line(-3,-2){ 30}}
\put(410,620){\line(-3,-2){ 30}}
\put(410,580){\line(-3,-2){ 30}}
\put(390,740){\line( 1,-1){ 20}}
\put(410,700){\line( 0,-1){ 20}}
\put(270,805){\makebox(0,0)[b]{\smash{\SetFigFont{10}{12.0}{rm}$\eta_1$}}}
\put(390,805){\makebox(0,0)[b]{\smash{\SetFigFont{10}{12.0}{rm}$\eta_2$}}}
\put(340,665){\makebox(0,0)[b]{\smash{\SetFigFont{10}{12.0}{rm}$B\stck{\tau_{n-1}}$}}}
\put(410,625){\makebox(0,0)[b]{\smash{\SetFigFont{10}{12.0}{rm}$C\stck{\tau_{n-1}}$}}}
\put(410,665){\makebox(0,0)[b]{\smash{\SetFigFont{10}{12.0}{rm}$C\stck{\tau_n}$}}}
\put(340,705){\makebox(0,0)[b]{\smash{\SetFigFont{10}{12.0}{rm}$B\stck{\tau_n}$}}}
\put(270,745){\makebox(0,0)[b]{\smash{\SetFigFont{10}{12.0}{rm}$A\stck{\tau_n}$}}}
\put(270,665){\makebox(0,0)[b]{\smash{\SetFigFont{10}{12.0}{rm}$A\stck{\tau_1}$}}}
\put(340,525){\makebox(0,0)[b]{\smash{\SetFigFont{10}{12.0}{rm}where
$\tau_1=\sigma_1$ and $\tau_{i+1}=\sigma_2(\tau_i)$}}}
\put(390,745){\makebox(0,0)[b]{\smash{\SetFigFont{10}{12.0}{rm}$S_2\stck{\sigma(\tau_n,\tau_n)}$}}}
\put(310,785){\makebox(0,0)[b]{\smash{\SetFigFont{10}{12.0}{rm}$S_1\stck{\sigma_0}$}}}
\put(270,705){\makebox(0,0)[b]{\smash{\SetFigFont{10}{12.0}{rm}$A\stck{\tau_{n-1}}$}}}
\put(410,705){\makebox(0,0)[b]{\smash{\SetFigFont{10}{12.0}{rm}$S_3\stck{\tau_n}$}}}
\put(410,585){\makebox(0,0)[b]{\smash{\SetFigFont{10}{12.0}{rm}$C\stck{\tau_1}$}}}
\put(340,625){\makebox(0,0)[b]{\smash{\SetFigFont{10}{12.0}{rm}$B\stck{\tau_1}$}}}
\end{picture}
\end{center}
\caption{Tree set for  $\setof{a^nb^nc^n}{$n\ge 1$}$}
\label{fig-pltg}
\end{figure}

This limitation can be overcome by moving to a formalism that
manipulates trees rather than stacks. We consider an extension of {\cfg}
in which each nonterminal $A$ is associated with a tree $\tau$.
Productions now specify how the tree associated with the mother is
related to the trees associated with the daughters. We denote trees with
first order terms. For example, the following production requires that
the $x$ and $y$ subtrees of the mother's tree are shared with the $B$
and $C$ daughters, respectively. In addition, the daughters have in
common the subtree $z$.

\[\begin{array}{ll}
A\tree{\sigma_0(x,y)}\rightarrow
& B\tree{\sigma_1(x,z)}\\
& C\tree{\sigma_2(y,z)}
\end{array}\]

There is a need to incorporate some kind of generalized notion of
linearity into such a system.  Corresponding to the {\em linearity\/}
restriction in {\lig} we require that any part of the mother's tree is
passed to at most one daughter. Corresponding to the {\em partial\/}
linearity of {\plig}, we permit subtrees that are not shared with the
mother to be shared amongst the daughters.  Under these conditions, the
tree set shown in Figure~\ref{fig-pltg} can be generated.  The nodes
$\eta_1$ and $\eta_2$ share the tree $\tau_n$, which occurs twice at the
node $\eta_2$. At $\eta_2$ the two copies of $\tau_n$ are distributed
across the daughters.

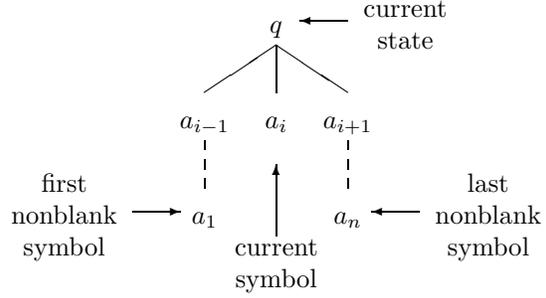
\begin{figure}
\setlength{\unitlength}{0.012500in}%
\begingroup\makeatletter\ifx\SetFigFont\undefined
% extract first six characters in \fmtname
\def\x#1#2#3#4#5#6#7\relax{\def\x{#1#2#3#4#5#6}}%
\expandafter\x\fmtname xxxxxx\relax \def\y{splain}%
\ifx\x\y   % LaTeX or SliTeX?
\gdef\SetFigFont#1#2#3{%
  \ifnum #1<17\tiny\else \ifnum #1<20\small\else
  \ifnum #1<24\normalsize\else \ifnum #1<29\large\else
  \ifnum #1<34\Large\else \ifnum #1<41\LARGE\else
     \huge\fi\fi\fi\fi\fi\fi
  \csname #3\endcsname}%
\else
\gdef\SetFigFont#1#2#3{\begingroup
  \count@#1\relax \ifnum 25<\count@\count@25\fi
  \def\x{\endgroup\@setsize\SetFigFont{#2pt}}%
  \expandafter\x
    \csname \romannumeral\the\count@ pt\expandafter\endcsname
    \csname @\romannumeral\the\count@ pt\endcsname
  \csname #3\endcsname}%
\fi
\fi\endgroup
\begin{picture}(177,113)(193,605)
\thinlines
\multiput(340,660)(0.00000,-8.00000){3}{\line( 0,-1){  4.000}}
\put(370,630){\vector(-1, 0){ 20}}
\put(340,665){\makebox(0,0)[b]{\smash{\SetFigFont{10}{12.0}{it}$a_{i+1}$}}}
\put(340,625){\makebox(0,0)[b]{\smash{\SetFigFont{10}{12.0}{rm}$a_n$}}}
\put(370,625){\makebox(0,0)[lb]{\smash{\SetFigFont{10}{12.0}{rm}\rightmostsym}}}
\put(310,680){\line( 0, 1){ 20}}
\put(310,620){\vector( 0, 1){ 30}}
\put(340,710){\vector(-1, 0){ 20}}
\put(280,680){\line( 3, 2){ 30}}
\put(310,700){\line( 3,-2){ 30}}
\multiput(280,660)(0.00000,-8.00000){3}{\line( 0,-1){  4.000}}
\put(250,630){\vector( 1, 0){ 20}}
\put(310,705){\makebox(0,0)[b]{\smash{\SetFigFont{10}{12.0}{rm}$q$}}}
\put(310,605){\makebox(0,0)[b]{\smash{\SetFigFont{10}{12.0}{rm}\mlines{current\\symbol}}}}
\put(340,705){\makebox(0,0)[lb]{\smash{\SetFigFont{10}{12.0}{rm}\mlines{current\\state}}}}
\put(280,665){\makebox(0,0)[b]{\smash{\SetFigFont{10}{12.0}{rm}$a_{i-1}$}}}
\put(280,625){\makebox(0,0)[b]{\smash{\SetFigFont{10}{12.0}{rm}$a_1$}}}
\put(250,625){\makebox(0,0)[rb]{\smash{\SetFigFont{10}{12.0}{rm}\leftmostsym}}}
\put(310,665){\makebox(0,0)[b]{\smash{\SetFigFont{10}{12.0}{rm}$a_i$}}}
\end{picture}
\smallskip
\caption{Encoding a Turing Machine}
\label{fig-tm}
\end{figure}

The formalism as currently described can be used to simulate arbitrary
Turing Machine computations. To see this, note that an instantaneous
description of a Turing Machine can be encoded with a tree as shown in
Figure~\ref{fig-tm}.  Moves of the Turing Machine can be simulated by
unary productions. The following production may be glossed: ``if in
state $q$ and scanning the symbol $X$, then change state to $q'$, write
the symbol $Y$ and move left'' \footnote{There will be a set of such
productions for each tape symbol $W$.}.

\[
A\tree{q(W(x),X,y)} \rightarrow A\tree{q'(x,W,Y(y))}
\]

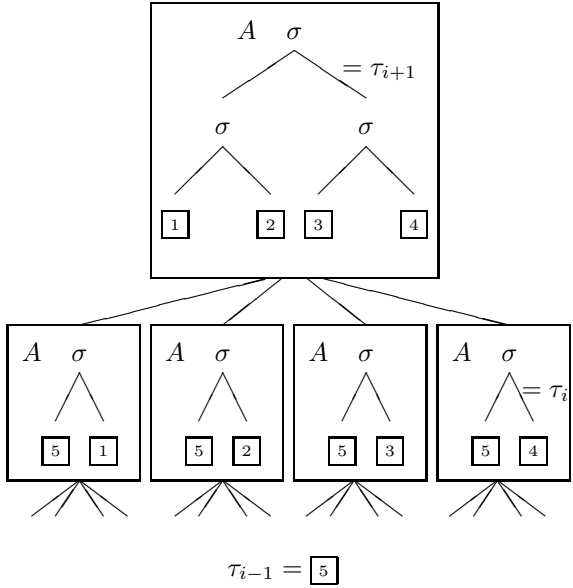
\begin{figure}
\begin{center}
\setlength{\unitlength}{0.012500in}%
\begingroup\makeatletter\ifx\SetFigFont\undefined
% extract first six characters in \fmtname
\def\x#1#2#3#4#5#6#7\relax{\def\x{#1#2#3#4#5#6}}%
\expandafter\x\fmtname xxxxxx\relax \def\y{splain}%
\ifx\x\y   % LaTeX or SliTeX?
\gdef\SetFigFont#1#2#3{%
  \ifnum #1<17\tiny\else \ifnum #1<20\small\else
  \ifnum #1<24\normalsize\else \ifnum #1<29\large\else
  \ifnum #1<34\Large\else \ifnum #1<41\LARGE\else
     \huge\fi\fi\fi\fi\fi\fi
  \csname #3\endcsname}%
\else
\gdef\SetFigFont#1#2#3{\begingroup
  \count@#1\relax \ifnum 25<\count@\count@25\fi
  \def\x{\endgroup\@setsize\SetFigFont{#2pt}}%
  \expandafter\x
    \csname \romannumeral\the\count@ pt\expandafter\endcsname
    \csname @\romannumeral\the\count@ pt\endcsname
  \csname #3\endcsname}%
\fi
\fi\endgroup
\begin{picture}(235,240)(210,595)
\thinlines
\put(240,635){\line(-4,-3){ 20}}
\put(240,635){\line(-2,-3){ 10}}
\put(240,635){\line( 2,-3){ 10}}
\put(240,635){\line( 4,-3){ 20}}
\put(300,635){\line(-4,-3){ 20}}
\put(300,635){\line(-2,-3){ 10}}
\put(300,635){\line( 2,-3){ 10}}
\put(300,635){\line( 4,-3){ 20}}
\put(355,635){\line(-4,-3){ 20}}
\put(355,635){\line(-2,-3){ 10}}
\put(355,635){\line( 2,-3){ 10}}
\put(355,635){\line( 4,-3){ 20}}
\put(415,635){\line(-4,-3){ 20}}
\put(415,635){\line(-2,-3){ 10}}
\put(415,635){\line( 2,-3){ 10}}
\put(415,635){\line( 4,-3){ 20}}
\put(240,700){\line( 4, 1){ 80}}
\put(340,720){\line( 4,-1){ 80}}
\put(335,720){\line( 5,-4){ 25}}
\put(325,720){\line(-5,-4){ 25}}
\put(230,660){\line( 1, 2){ 10}}
\put(240,680){\line( 1,-2){ 10}}
\put(290,660){\line( 1, 2){ 10}}
\put(300,680){\line( 1,-2){ 10}}
\put(350,660){\line( 1, 2){ 10}}
\put(360,680){\line( 1,-2){ 10}}
\put(410,660){\line( 1, 2){ 10}}
\put(420,680){\line( 1,-2){ 10}}
\put(330,635){\framebox(55,65){}}
\put(210,635){\framebox(55,65){}}
\put(270,635){\framebox(55,65){}}
\put(390,635){\framebox(55,65){}}
\put(270,720){\framebox(120,115){}}
\put(300,795){\line( 3, 2){ 30}}
\put(330,815){\line( 3,-2){ 30}}
\put(300,775){\line(-1,-1){ 20}}
\put(300,775){\line( 1,-1){ 20}}
\put(360,775){\line(-1,-1){ 20}}
\put(360,775){\line( 1,-1){ 20}}
\put(240,685){\makebox(0,0)[b]{\smash{\SetFigFont{10}{12.0}{rm}$\sigma$}}}
\put(360,685){\makebox(0,0)[b]{\smash{\SetFigFont{10}{12.0}{rm}$\sigma$}}}
\put(420,685){\makebox(0,0)[b]{\smash{\SetFigFont{10}{12.0}{rm}$\sigma$}}}
\put(220,685){\makebox(0,0)[b]{\smash{\SetFigFont{10}{12.0}{rm}$A$}}}
\put(280,685){\makebox(0,0)[b]{\smash{\SetFigFont{10}{12.0}{rm}$A$}}}
\put(340,685){\makebox(0,0)[b]{\smash{\SetFigFont{10}{12.0}{rm}$A$}}}
\put(400,685){\makebox(0,0)[b]{\smash{\SetFigFont{10}{12.0}{rm}$A$}}}
\put(230,645){\makebox(0,0)[b]{\smash{\SetFigFont{6}{7.2}{rm}\fbox{5}}}}
\put(300,685){\makebox(0,0)[b]{\smash{\SetFigFont{10}{12.0}{rm}$\sigma$}}}
\put(290,645){\makebox(0,0)[b]{\smash{\SetFigFont{6}{7.2}{rm}\fbox{5}}}}
\put(350,645){\makebox(0,0)[b]{\smash{\SetFigFont{6}{7.2}{rm}\fbox{5}}}}
\put(410,645){\makebox(0,0)[b]{\smash{\SetFigFont{6}{7.2}{rm}\fbox{5}}}}
\put(250,645){\makebox(0,0)[b]{\smash{\SetFigFont{6}{7.2}{rm}\fbox{1}}}}
\put(310,645){\makebox(0,0)[b]{\smash{\SetFigFont{6}{7.2}{rm}\fbox{2}}}}
\put(370,645){\makebox(0,0)[b]{\smash{\SetFigFont{6}{7.2}{rm}\fbox{3}}}}
\put(430,645){\makebox(0,0)[b]{\smash{\SetFigFont{6}{7.2}{rm}\fbox{4}}}}
\put(425,670){\makebox(0,0)[lb]{\smash{\SetFigFont{10}{12.0}{rm}$=\tau_i$}}}
\put(360,780){\makebox(0,0)[b]{\smash{\SetFigFont{10}{12.0}{rm}$\sigma$}}}
\put(300,780){\makebox(0,0)[b]{\smash{\SetFigFont{10}{12.0}{rm}$\sigma$}}}
\put(310,820){\makebox(0,0)[b]{\smash{\SetFigFont{10}{12.0}{rm}$A$}}}
\put(330,820){\makebox(0,0)[b]{\smash{\SetFigFont{10}{12.0}{rm}$\sigma$}}}
\put(350,805){\makebox(0,0)[lb]{\smash{\SetFigFont{10}{12.0}{rm}$=\tau_{i+1}$}}}
\put(280,740){\makebox(0,0)[b]{\smash{\SetFigFont{6}{7.2}{rm}\fbox{1}}}}
\put(320,740){\makebox(0,0)[b]{\smash{\SetFigFont{6}{7.2}{rm}\fbox{2}}}}
\put(340,740){\makebox(0,0)[b]{\smash{\SetFigFont{6}{7.2}{rm}\fbox{3}}}}
\put(380,740){\makebox(0,0)[b]{\smash{\SetFigFont{6}{7.2}{rm}\fbox{4}}}}
\put(325,595){\makebox(0,0)[b]{\smash{\SetFigFont{10}{12.0}{rm}$\tau_{i-1}=\fbox{\tiny 5}$}}}
\end{picture}
\end{center}
\caption{Building full binary trees}
\label{fig-fbinary}
\end{figure}

One solution to this problem is to prevent a single daughter sharing
more than one of its subtrees with the mother. However, we do not impose
this restriction because it still leaves open the possibility of
generating trees in which every branch has the same length, thus
violating the condition that trees have at most a bounded number of
unbounded, dependent branches.  Figure~\ref{fig-fbinary} shows how a set
of such trees can be generated by illustrating the effect of the
following production.

\[\begin{array}{ll}
A\tree{\sigma(\sigma(x,y),\sigma(x',y'))} \rightarrow &
A\tree{\sigma(z,x)}\\&A\tree{\sigma(z,y)}\\
&A\tree{\sigma(z,x')}\\&A\tree{\sigma(z,y')}
\end{array}\]

To see this, assume (by induction) that all four of the daughter
nonterminals are associated with the full binary tree of height $i$
($\tau_i$).  All four of these trees are constrained to be equal by the
production given above, which requires that they have identical left
(i.e. $z$) subtrees (these subtrees must be the full binary tree
$\tau_{i-1}$).  Passing the right subtrees ($x$, $y$, $x'$ and $y'$) to
the mother as shown allows the construction of a full binary tree with
height $i+1$ ($\tau_{i+1}$). This can be repeated an unbounded
number of times so that all full binary trees are produced.

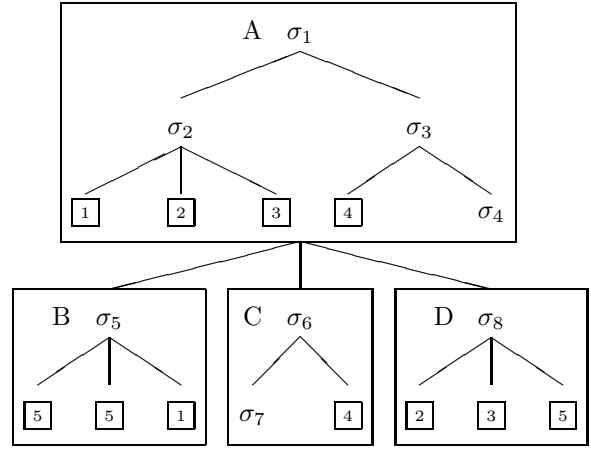
\begin{figure}
\begin{center}
\setlength{\unitlength}{0.012500in}%
\begingroup\makeatletter\ifx\SetFigFont\undefined
% extract first six characters in \fmtname
\def\x#1#2#3#4#5#6#7\relax{\def\x{#1#2#3#4#5#6}}%
\expandafter\x\fmtname xxxxxx\relax \def\y{splain}%
\ifx\x\y   % LaTeX or SliTeX?
\gdef\SetFigFont#1#2#3{%
  \ifnum #1<17\tiny\else \ifnum #1<20\small\else
  \ifnum #1<24\normalsize\else \ifnum #1<29\large\else
  \ifnum #1<34\Large\else \ifnum #1<41\LARGE\else
     \huge\fi\fi\fi\fi\fi\fi
  \csname #3\endcsname}%
\else
\gdef\SetFigFont#1#2#3{\begingroup
  \count@#1\relax \ifnum 25<\count@\count@25\fi
  \def\x{\endgroup\@setsize\SetFigFont{#2pt}}%
  \expandafter\x
    \csname \romannumeral\the\count@ pt\expandafter\endcsname
    \csname @\romannumeral\the\count@ pt\endcsname
  \csname #3\endcsname}%
\fi
\fi\endgroup
\begin{picture}(248,185)(216,535)
\thinlines
\put(250,640){\line( 2, 1){ 40}}
\put(290,660){\line( 2,-1){ 40}}
\put(290,680){\line( 5, 2){ 50}}
\put(340,700){\line( 5,-2){ 50}}
\put(290,660){\line( 0,-1){ 20}}
\put(390,660){\line(-3,-2){ 30}}
\put(390,660){\line( 3,-2){ 30}}
\put(240,620){\framebox(190,100){}}
\put(320,705){\makebox(0,0)[b]{\smash{\SetFigFont{10}{12.0}{rm}A}}}
\put(250,630){\makebox(0,0)[b]{\smash{\SetFigFont{6}{7.2}{rm}\fbox{1}}}}
\put(290,630){\makebox(0,0)[b]{\smash{\SetFigFont{6}{7.2}{rm}\fbox{2}}}}
\put(330,630){\makebox(0,0)[b]{\smash{\SetFigFont{6}{7.2}{rm}\fbox{3}}}}
\put(360,630){\makebox(0,0)[b]{\smash{\SetFigFont{6}{7.2}{rm}\fbox{4}}}}
\put(340,705){\makebox(0,0)[b]{\smash{\SetFigFont{10}{12.0}{rm}$\sigma_1$}}}
\put(290,665){\makebox(0,0)[b]{\smash{\SetFigFont{10}{12.0}{rm}$\sigma_2$}}}
\put(390,665){\makebox(0,0)[b]{\smash{\SetFigFont{10}{12.0}{rm}$\sigma_3$}}}
\put(420,630){\makebox(0,0)[b]{\smash{\SetFigFont{10}{12.0}{rm}$\sigma_4$}}}
\put(320,560){\line( 1, 1){ 20}}
\put(340,580){\line( 1,-1){ 20}}
\put(420,580){\line( 0,-1){ 20}}
\put(340,620){\line(-4,-1){ 80}}
\put(340,620){\line( 0,-1){ 20}}
\put(340,620){\line( 4,-1){ 80}}
\put(260,580){\line( 0,-1){ 20}}
\put(260,580){\line(-3,-2){ 30}}
\put(260,580){\line( 3,-2){ 30}}
\put(420,580){\line(-3,-2){ 30}}
\put(420,580){\line( 3,-2){ 30}}
\put(220,535){\framebox(80,65){}}
\put(310,535){\framebox(60,65){}}
\put(380,535){\framebox(80,65){}}
\put(240,585){\makebox(0,0)[b]{\smash{\SetFigFont{10}{12.0}{rm}B}}}
\put(320,585){\makebox(0,0)[b]{\smash{\SetFigFont{10}{12.0}{rm}C}}}
\put(400,585){\makebox(0,0)[b]{\smash{\SetFigFont{10}{12.0}{rm}D}}}
\put(260,545){\makebox(0,0)[b]{\smash{\SetFigFont{6}{7.2}{rm}\fbox{5}}}}
\put(260,585){\makebox(0,0)[b]{\smash{\SetFigFont{10}{12.0}{rm}$\sigma_5$}}}
\put(340,585){\makebox(0,0)[b]{\smash{\SetFigFont{10}{12.0}{rm}$\sigma_6$}}}
\put(420,585){\makebox(0,0)[b]{\smash{\SetFigFont{10}{12.0}{rm}$\sigma_8$}}}
\put(320,545){\makebox(0,0)[b]{\smash{\SetFigFont{10}{12.0}{rm}$\sigma_7$}}}
\put(360,545){\makebox(0,0)[b]{\smash{\SetFigFont{6}{7.2}{rm}\fbox{4}}}}
\put(420,545){\makebox(0,0)[b]{\smash{\SetFigFont{6}{7.2}{rm}\fbox{3}}}}
\put(230,545){\makebox(0,0)[b]{\smash{\SetFigFont{6}{7.2}{rm}\fbox{5}}}}
\put(290,545){\makebox(0,0)[b]{\smash{\SetFigFont{6}{7.2}{rm}\fbox{1}}}}
\put(390,545){\makebox(0,0)[b]{\smash{\SetFigFont{6}{7.2}{rm}\fbox{2}}}}
\put(450,545){\makebox(0,0)[b]{\smash{\SetFigFont{6}{7.2}{rm}\fbox{5}}}}
\end{picture}
\end{center}
\caption{A \pltg\ local tree}
\label{fig-pltga}
\end{figure}

To overcome both of these problems we impose the following additional
constraint on the productions of a grammar. We require that subtrees of
the mother that are passed to daughters that share subtrees with one
another must appear as siblings in the mother's tree.  Note that this
condition rules out the production responsible for building full binary
trees since the $x, y, x'$ and $y'$ subtrees are not siblings in the
mother's tree despite the fact that all of the daughters share a common
subtree $z$.  Moreover, since a daughter shares subtrees with itself, a
special case of the condition is that subtrees occurring within some
daughter can only appear as siblings in the mother.  This condition also
rules out the Turing Machine simulation.  We refer to this formalism as
Partially Linear Tree Grammars (\pltg). As a further illustration of the
constraints places on shared subtrees, Figure~\ref{fig-pltga} shows a
local tree that could appear in a derivation tree. This local tree is
licensed by the following production which respects all of the
constraints on \pltg\ productions.

\[\begin{array}{l}
A\tree{\sigma_1(\sigma_2(x_1,x_2,x_3),\sigma_3(x_4,\sigma_4))}
\rightarrow \\
\qquad B\tree{\sigma_5(x_5,x_5,x_1)}\\
\qquad C\tree{\sigma_6(\sigma_7,x_4)}\\
\qquad D\tree{\sigma_8(x_2,x_3,x_5)}
\end{array}\]
Note that in Figure~\ref{fig-pltga} the daughter nodes labelled B and D share
a common subtree and the subtrees shared between the mother and the B and D
daughters appear as siblings in the tree associated with the mother.

\begin{Example}The \pltg\ with the following productions  generates
the language $$\setof{a^nb^nc^n}{$n\ge 1$}$$ and the tree set
shown in Figure~\ref{fig-pltg}.
\[
\begin{array}[c]{ll}
S_1\tree{\sigma_0}\rightarrow A\tree{x}\,S_2\tree{\sigma(x,x)},
\\
S_2\tree{\sigma(x,y)}\rightarrow B\tree{x}\,S_3\tree{y}, \\
S_3\tree{x} \rightarrow C\tree{x}, & \\
A\tree{\sigma_2(x)}\rightarrow aA\tree{x}, \qquad\qquad &
A\tree{\sigma_1}\rightarrow a, \\
B\tree{\sigma_2(x)}\rightarrow bB\tree{x}, &
B\tree{\sigma_1}\rightarrow b, \\
C\tree{\sigma_2(x)}\rightarrow cC\tree{x}, &
C\tree{\sigma_1}\rightarrow c.
\end{array}
\]
\end{Example}

\begin{Example}
The \pltg\ with the following productions generates the language
of strings consisting of $k$ copies of strings of matching parenthesis,
i.e., the language
\[\setof{w^k}{$w\in D$}\] where $k\ge 1$ and
$D$ is the set of strings in $\set{(,)}^*$ that have balanced brackets,
i.e, the Dyck language over $\set{(,)}$.
\[
\begin{array}[c]{ll}
S\stck{}\rightarrow
\underbrace{A\stck{x}\ldots A\stck{x}}_{\mbox{$k$ copies}}, &
A\stck{}\rightarrow\lambda,\\\\
A\stck{\sigma_1(x)}\rightarrow (\,A\stck{x}\,), &
A\stck{\sigma_2(x,y)}\rightarrow A\stck{x}\,A\stck{y}.
\end{array}
\]
\label{ex-kdyckcopy}
\end{Example}

\section{Trees to Feature Structures}

Finally, we note that acyclic feature structures without re-entrancy can
be viewed as trees with branches labelled by feature names and atomic
values only found at leaf nodes (interior nodes being unlabelled). Based
on this observation, we can consider the constraints we have formulated
for the tree system {\pltg} as constraints on a unification-based
grammar formalism such as {\PATR}. We will call this system Partially
Linear \PATR\ (\plpatr).  Having made the move from trees to feature
structures, we consider the possibility of re-entrancy in {\plpatr}.

Note that the feature structure at the root of a \plpatr\ derivation
tree will not involve re-entrancy.  However, for the following reasons
we believe that this does not constitute as great a limitation as it
might appear.  In unification-based grammar, the feature structure
associated with the root of the tree is often regarded as the structure
that has been derived from the input (i.e., the output of a parser). As
a consequence there is a tendency to use the grammar rules to accumulate
a single, large feature structure giving a complete encoding of the
analysis.  To do this, unbounded feature information is passed up the
tree in a way that violates the constraints developed in this paper.
Rather than giving such prominence to the root feature structure, we
suggest that the entire derivation tree should be seen as the object
that is derived from the input, i.e., this is what the parser returns.
Because feature structures associated with all nodes in the tree are
available, feature information need only be passed up the tree when it
is required in order to establish dependencies within the derivation
tree. When this approach is taken, there may be less need for
re-entrancy in the root feature structure. Furthermore, re-entrancy in
the form of shared feature structures within and across nodes will be
found in \plpatr\ (see for example Figure~\ref{fig-pltga}).

\section{Generative Capacity}

\lig\ are more powerful than \cfg\ and are known to be weakly equivalent
to Tree Adjoining Grammar, Combinatory Categorial Grammar, and Head
Grammar~\cite{vw94}. \plig\ are more powerful than \lig\ since they can
generate the $k$-copy language for any fixed $k$ (see
Example~\ref{ex-kcopy}).  Slightly more generally, \plig\ can generate
the language \[\setof{w^k}{$w\in R$}\] for any $k\ge 1$ and regular
language $R$.  We believe that the language involving copies of strings
of matching brackets described in Example~\ref{ex-kdyckcopy} cannot be
generated by \plig\ but, as shown in Example~\ref{ex-kdyckcopy}, it can
be generated by \pltg\ and therefore \plpatr.  Slightly more generally,
\pltg\ can generate the language \[\setof{w^k}{$w\in L$}\] for any $k\ge
1$ and context-free language $L$. It appears that the class of languages
generated by \pltg\ is included in those languages generated by Linear
Context-Free Rewriting Systems~\cite{vwj87b} since the construction
involved in a proof of this underlies the recognition algorithm
discussed in the next section.

As is the case for the tree sets of \ig, \lig\ and Tree
Adjoining Grammar, the tree sets generated by \pltg\ have path sets that
are context-free languages. In other words, the set of all strings
labelling root to frontier paths of derivation trees is a context-free
language.  While the tree sets of \lig\ and Tree Adjoining Grammars have
independent branches, \pltg\ tree sets exhibit dependent branches,
where the number of dependent branches in any tree is bounded by the
grammar. Note that the number of dependent branches in the tree sets of
\ig\ is not bounded by the grammar (e.g., they generate
sets of all full binary trees).

\section{Tractable Recognition}

In this section we outline the main ideas underlying a polynomial time
recognition algorithm for \plpatr\ that generalizes the \cky\ algorithm
\cite{ka65,y67}.  The key to this algorithm is the use of structure
sharing techniques similar to those used to process \lig\ efficiently
\cite{vw93b}. To understand how these techniques are applied in the case
of \plpatr, it is therefore helpful to consider first the somewhat
simpler case of \lig.

The \cky\ algorithm is a bottom-up recognition algorithm for \cfg. For a
given grammar $G$ and input string $a_1\ldots a_n$ the algorithm
constructs an array $P$, having $n^2$ elements, where element $P[i,j]$
stores all and only those nonterminals of $G$ that derive the substring
$a_i \ldots a_j$. A naive adaptation of this algorithm for \lig\
recognition would involve storing a set of nonterminals and their
associated stacks. But since stack length is at least proportional to
the length of the input string, the resultant algorithm would exhibit
exponential space and time complexity in the worst case.  Vijay-Shanker
and Weir~\shortcite{vw93b} showed that the behaviour of the naive
algorithm can be improved upon.  In \lig\ derivations the application of
a rule cannot depend on more than a bounded portion of the top of the
stack. Thus, rather than storing the whole of the potentially unbounded
stack in a particular array entry, it suffices to store just a bounded
portion together with a pointer to the residue.

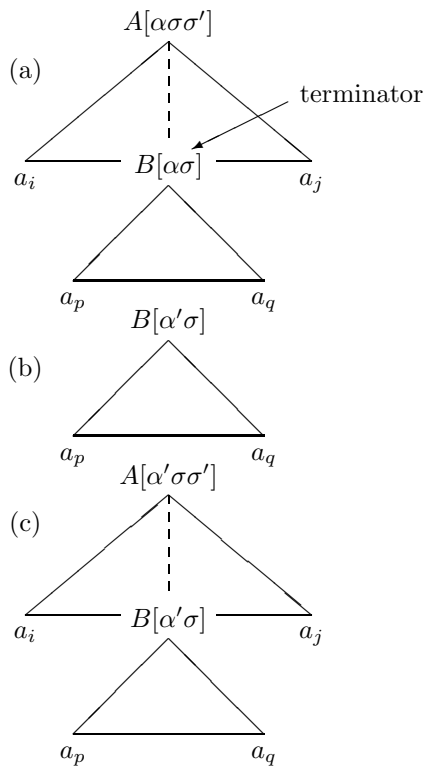
\begin{figure}
\begin{center}
\setlength{\unitlength}{0.012500in}%
\begingroup\makeatletter\ifx\SetFigFont\undefined
% extract first six characters in \fmtname
\def\x#1#2#3#4#5#6#7\relax{\def\x{#1#2#3#4#5#6}}%
\expandafter\x\fmtname xxxxxx\relax \def\y{splain}%
\ifx\x\y   % LaTeX or SliTeX?
\gdef\SetFigFont#1#2#3{%
  \ifnum #1<17\tiny\else \ifnum #1<20\small\else
  \ifnum #1<24\normalsize\else \ifnum #1<29\large\else
  \ifnum #1<34\Large\else \ifnum #1<41\LARGE\else
     \huge\fi\fi\fi\fi\fi\fi
  \csname #3\endcsname}%
\else
\gdef\SetFigFont#1#2#3{\begingroup
  \count@#1\relax \ifnum 25<\count@\count@25\fi
  \def\x{\endgroup\@setsize\SetFigFont{#2pt}}%
  \expandafter\x
    \csname \romannumeral\the\count@ pt\expandafter\endcsname
    \csname @\romannumeral\the\count@ pt\endcsname
  \csname #3\endcsname}%
\fi
\fi\endgroup
\begin{picture}(142,318)(89,420)
\thinlines
\put(140,670){\line(-1, 0){ 40}}
\put(100,670){\line( 6, 5){ 60}}
\put(160,720){\line( 6,-5){ 60}}
\put(220,670){\line(-1, 0){ 40}}
\multiput(160,720)(0.00000,-7.27273){6}{\line( 0,-1){  3.636}}
\put(160,660){\line(-1,-1){ 40}}
\put(120,620){\line( 1, 0){ 80}}
\put(200,620){\line(-1, 1){ 40}}
\put(210,695){\vector(-2,-1){ 40}}
\put(140,480){\line(-1, 0){ 40}}
\put(100,480){\line( 6, 5){ 60}}
\put(160,530){\line( 6,-5){ 60}}
\put(220,480){\line(-1, 0){ 40}}
\multiput(160,530)(0.00000,-7.27273){6}{\line( 0,-1){  3.636}}
\put(160,470){\line(-1,-1){ 40}}
\put(120,430){\line( 1, 0){ 80}}
\put(200,430){\line(-1, 1){ 40}}
\put(160,595){\line(-1,-1){ 40}}
\put(120,555){\line( 1, 0){ 80}}
\put(200,555){\line(-1, 1){ 40}}
\put(160,725){\makebox(0,0)[b]{\smash{\SetFigFont{10}{12.0}{rm}$A\stck{\alpha\sigma\sigma'}$}}}
\put(100,705){\makebox(0,0)[b]{\smash{\SetFigFont{10}{12.0}{rm}(a)}}}
\put(160,665){\makebox(0,0)[b]{\smash{\SetFigFont{10}{12.0}{rm}$B\stck{\alpha\sigma}$}}}
\put(215,695){\makebox(0,0)[lb]{\smash{\SetFigFont{10}{12.0}{rm}terminator}}}
\put(100,660){\makebox(0,0)[b]{\smash{\SetFigFont{10}{12.0}{rm}$a_i$}}}
\put(220,660){\makebox(0,0)[b]{\smash{\SetFigFont{10}{12.0}{rm}$a_j$}}}
\put(120,610){\makebox(0,0)[b]{\smash{\SetFigFont{10}{12.0}{rm}$a_p$}}}
\put(200,610){\makebox(0,0)[b]{\smash{\SetFigFont{10}{12.0}{rm}$a_q$}}}
\put(160,535){\makebox(0,0)[b]{\smash{\SetFigFont{10}{12.0}{rm}$A\stck{\alpha'\sigma\sigma'}$}}}
\put(160,475){\makebox(0,0)[b]{\smash{\SetFigFont{10}{12.0}{rm}$B\stck{\alpha'\sigma}$}}}
\put(100,515){\makebox(0,0)[b]{\smash{\SetFigFont{10}{12.0}{rm}(c)}}}
\put(160,600){\makebox(0,0)[b]{\smash{\SetFigFont{10}{12.0}{rm}$B\stck{\alpha'\sigma}$}}}
\put(100,580){\makebox(0,0)[b]{\smash{\SetFigFont{10}{12.0}{rm}(b)}}}
\put(120,545){\makebox(0,0)[b]{\smash{\SetFigFont{10}{12.0}{rm}$a_p$}}}
\put(200,545){\makebox(0,0)[b]{\smash{\SetFigFont{10}{12.0}{rm}$a_q$}}}
\put(100,470){\makebox(0,0)[b]{\smash{\SetFigFont{10}{12.0}{rm}$a_i$}}}
\put(220,470){\makebox(0,0)[b]{\smash{\SetFigFont{10}{12.0}{rm}$a_j$}}}
\put(120,420){\makebox(0,0)[b]{\smash{\SetFigFont{10}{12.0}{rm}$a_p$}}}
\put(200,420){\makebox(0,0)[b]{\smash{\SetFigFont{10}{12.0}{rm}$a_q$}}}
\end{picture}
\end{center}
\caption{``Context-Freeness' in \lig\ derivations}
\label{fig-ligterm}
\end{figure}

Consider Figure~\ref{fig-ligterm}. Tree~(a) shows a \lig\ derivation of
the substring $a_i \ldots a_j$ from the object
$A[\alpha\sigma\sigma']$. In this derivation tree, the node labelled
$B[\alpha\sigma]$ is a distinguished descendant of the
root\protect{\footnote{The stack $\alpha\sigma$ associated with $B$ is
``inherited'' from the stack associated with $A$ at the root of the
tree.}} and is the first point below $A[\alpha\sigma\sigma']$ at which
the top symbol ($\sigma$) of the (unbounded) stack $\alpha\sigma$ is
exposed. This node is called the {\em terminator\/} of the node labelled
$A[\alpha\sigma]$.  It is not difficult to show that only that portion
of the derivation {\em below\/} the terminator node is dependent on more
than the top of the stack $\alpha\sigma$. It follows that for any stack
$\alpha'\sigma$, if there is a derivation of the substring $a_p \ldots
a_q$ from $B[\alpha'\sigma]$ (see tree~(b)), then there is a
corresponding derivation of $a_i \ldots a_j$ from
$A[\alpha'\sigma\sigma']$ (see tree~(c)). This captures the sense in
which \lig\ derivations exhibit ``context-freeness''.  Efficient storage of
stacks can therefore be achieved by storing in $P[i,j]$ just that
bounded amount of information (nonterminal plus top of stack) relevant
to rule application, together with a pointer to any entry in $P[p,q]$
representing a subderivation from an object $B[\alpha'\sigma]$.

Before describing how we adapt this technique to the case of \plpatr\ we
discuss the sense in which \plpatr\ derivations exhibit a
``context-freeness'' property.  The constraints on \plpatr\ which we
have identified in this paper ensure that these feature values can be
manipulated independently of one another and that they behave in a
stack-like way.  As a consequence, the storage technique used
effectively for \lig\ recognition may be generalized to the case of
\plpatr.

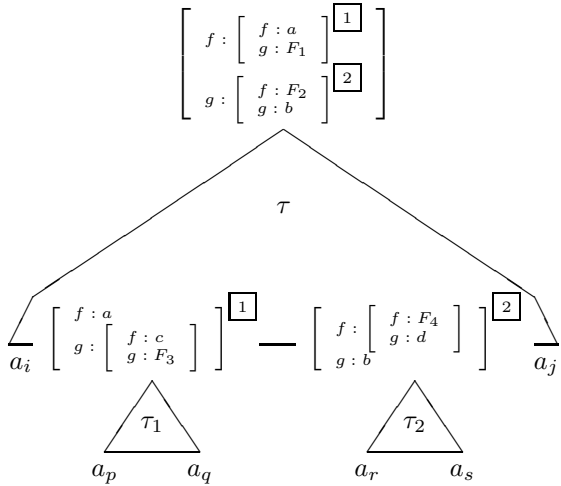
\begin{figure}
\begin{center}
\setlength{\unitlength}{0.012500in}%
\begingroup\makeatletter\ifx\SetFigFont\undefined
% extract first six characters in \fmtname
\def\x#1#2#3#4#5#6#7\relax{\def\x{#1#2#3#4#5#6}}%
\expandafter\x\fmtname xxxxxx\relax \def\y{splain}%
\ifx\x\y   % LaTeX or SliTeX?
\gdef\SetFigFont#1#2#3{%
  \ifnum #1<17\tiny\else \ifnum #1<20\small\else
  \ifnum #1<24\normalsize\else \ifnum #1<29\large\else
  \ifnum #1<34\Large\else \ifnum #1<41\LARGE\else
     \huge\fi\fi\fi\fi\fi\fi
  \csname #3\endcsname}%
\else
\gdef\SetFigFont#1#2#3{\begingroup
  \count@#1\relax \ifnum 25<\count@\count@25\fi
  \def\x{\endgroup\@setsize\SetFigFont{#2pt}}%
  \expandafter\x
    \csname \romannumeral\the\count@ pt\expandafter\endcsname
    \csname @\romannumeral\the\count@ pt\endcsname
  \csname #3\endcsname}%
\fi
\fi\endgroup
\begin{picture}(242,183)(179,495)
\thinlines
\put(245,535){\line(-2,-3){ 20}}
\put(225,505){\line( 1, 0){ 40}}
\put(265,505){\line(-2, 3){ 20}}
\put(245,550){\makebox(0,0)[b]{\smash{\SetFigFont{10}{12.0}{rm}$\ftermfs$}}}
\put(225,495){\makebox(0,0)[b]{\smash{\SetFigFont{10}{12.0}{rm}$a_p$}}}
\put(265,495){\makebox(0,0)[b]{\smash{\SetFigFont{10}{12.0}{rm}$a_q$}}}
\put(245,515){\makebox(0,0)[b]{\smash{\SetFigFont{10}{12.0}{rm}$\tau_1$}}}
\put(355,535){\line(-2,-3){ 20}}
\put(335,505){\line( 1, 0){ 40}}
\put(375,505){\line(-2, 3){ 20}}
\put(355,550){\makebox(0,0)[b]{\smash{\SetFigFont{10}{12.0}{rm}$\gtermfs$}}}
\put(335,495){\makebox(0,0)[b]{\smash{\SetFigFont{10}{12.0}{rm}$a_r$}}}
\put(375,495){\makebox(0,0)[b]{\smash{\SetFigFont{10}{12.0}{rm}$a_s$}}}
\put(355,515){\makebox(0,0)[b]{\smash{\SetFigFont{10}{12.0}{rm}$\tau_2$}}}
\put(290,550){\line( 1, 0){ 15}}
\put(300,640){\line(-3,-2){105}}
\put(195,570){\line(-1,-2){ 10}}
\put(185,550){\line( 1, 0){ 10}}
\put(300,640){\line( 3,-2){105}}
\put(405,570){\line( 1,-2){ 10}}
\put(415,550){\line(-1, 0){ 10}}
\put(300,665){\makebox(0,0)[b]{\smash{\SetFigFont{10}{12.0}{rm}$\rootfs$}}}
\put(300,605){\makebox(0,0)[b]{\smash{\SetFigFont{10}{12.0}{rm}$\tau$}}}
\put(190,540){\makebox(0,0)[b]{\smash{\SetFigFont{10}{12.0}{rm}$a_i$}}}
\put(410,540){\makebox(0,0)[b]{\smash{\SetFigFont{10}{12.0}{rm}$a_j$}}}
\end{picture}
\end{center}
\caption{Terminators in \plpatr}
\label{fig-reca}
\end{figure}

Suppose that we have the derived tree shown in Figure~\ref{fig-reca}
where the nodes at the root of the subtrees $\tau_1$ and $\tau_2$ are
the so-called $f$-terminator and $g$-terminator of the tree's root,
respectively. Roughly speaking, the $f$-terminator of a node is the node
from which it gets the value for the feature $f$. Because of the
constraints on the form of \plpatr\ productions, the derivations between
the root of $\tau$ and these terminators cannot in general depend on
more than a bounded part of the feature structures $\boxed{1}$ and
$\boxed{2}$. At the root of the figure the feature structures
$\boxed{1}$ and $\boxed{2}$ have been expanded to show the extent of the
dependency in this example. In this case, the value of the feature $f$
in $\boxed{1}$ must be $a$, whereas, the feature $g$ is not
fixed. Furthermore, the value of the feature $g$ in $\boxed{2}$ must be
$b$, whereas, the feature $f$ is not fixed.  This means, for example,
that the applicability of the productions used on the path from the root
of $\tau_1$ to the root of $\tau$ depends on the feature $f$ in
$\boxed{1}$ having the value $a$ but does not depend on the value of the
feature $g$ in $\boxed{1}$.  Note that in this tree the value of the
feature $g$ in $\boxed{1}$ is
\[F_1=\fsnorm{f:c\\g:F_3}\] and the value of the feature $f$ in $\boxed{2}$
is \[F_2=\fsnorm{f:F_4\\g:d}\]

\begin{figure}
\begin{center}
\setlength{\unitlength}{0.012500in}%
\begingroup\makeatletter\ifx\SetFigFont\undefined
% extract first six characters in \fmtname
\def\x#1#2#3#4#5#6#7\relax{\def\x{#1#2#3#4#5#6}}%
\expandafter\x\fmtname xxxxxx\relax \def\y{splain}%
\ifx\x\y   % LaTeX or SliTeX?
\gdef\SetFigFont#1#2#3{%
  \ifnum #1<17\tiny\else \ifnum #1<20\small\else
  \ifnum #1<24\normalsize\else \ifnum #1<29\large\else
  \ifnum #1<34\Large\else \ifnum #1<41\LARGE\else
     \huge\fi\fi\fi\fi\fi\fi
  \csname #3\endcsname}%
\else
\gdef\SetFigFont#1#2#3{\begingroup
  \count@#1\relax \ifnum 25<\count@\count@25\fi
  \def\x{\endgroup\@setsize\SetFigFont{#2pt}}%
  \expandafter\x
    \csname \romannumeral\the\count@ pt\expandafter\endcsname
    \csname @\romannumeral\the\count@ pt\endcsname
  \csname #3\endcsname}%
\fi
\fi\endgroup
\begin{picture}(242,183)(179,495)
\thinlines
\put(290,550){\line( 1, 0){ 15}}
\put(300,640){\line(-3,-2){105}}
\put(195,570){\line(-1,-2){ 10}}
\put(185,550){\line( 1, 0){ 10}}
\put(300,640){\line( 3,-2){105}}
\put(405,570){\line( 1,-2){ 10}}
\put(415,550){\line(-1, 0){ 10}}
\put(245,535){\line(-2,-3){ 20}}
\put(225,505){\line( 1, 0){ 40}}
\put(265,505){\line(-2, 3){ 20}}
\put(355,535){\line(-2,-3){ 20}}
\put(335,505){\line( 1, 0){ 40}}
\put(375,505){\line(-2, 3){ 20}}
\put(190,540){\makebox(0,0)[b]{\smash{\SetFigFont{10}{12.0}{rm}$a_i$}}}
\put(410,540){\makebox(0,0)[b]{\smash{\SetFigFont{10}{12.0}{rm}$a_j$}}}
\put(300,665){\makebox(0,0)[b]{\smash{\SetFigFont{10}{12.0}{rm}$\rootfsb$}}}
\put(225,495){\makebox(0,0)[b]{\smash{\SetFigFont{10}{12.0}{rm}$a_p$}}}
\put(265,495){\makebox(0,0)[b]{\smash{\SetFigFont{10}{12.0}{rm}$a_q$}}}
\put(335,495){\makebox(0,0)[b]{\smash{\SetFigFont{10}{12.0}{rm}$a_r$}}}
\put(375,495){\makebox(0,0)[b]{\smash{\SetFigFont{10}{12.0}{rm}$a_s$}}}
\put(245,550){\makebox(0,0)[b]{\smash{\SetFigFont{10}{12.0}{rm}$\ftermfsb$}}}
\put(355,550){\makebox(0,0)[b]{\smash{\SetFigFont{10}{12.0}{rm}$\gtermfsb$}}}
\put(245,515){\makebox(0,0)[b]{\smash{\SetFigFont{10}{12.0}{rm}$\tau_3$}}}
\put(355,515){\makebox(0,0)[b]{\smash{\SetFigFont{10}{12.0}{rm}$\tau_4$}}}
\put(300,605){\makebox(0,0)[b]{\smash{\SetFigFont{10}{12.0}{rm}$\tau'$}}}
\end{picture}
\end{center}
\caption{Alternative derivation}
\label{fig-recc}
\end{figure}
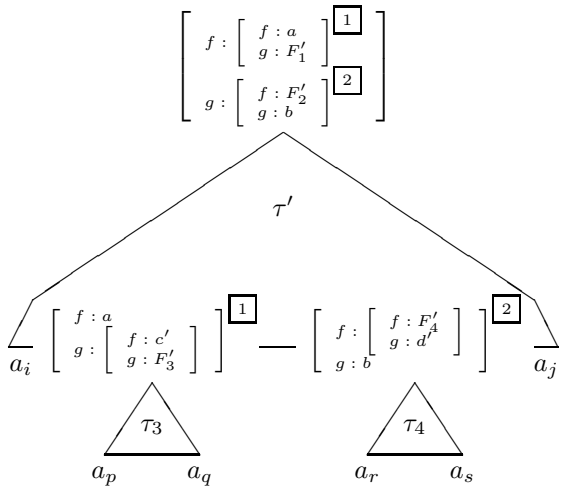

Suppose that, in addition to the tree shown in Figure~\ref{fig-reca} the
grammar generates the pair of trees shown in
Figure~\ref{fig-recb}. Notice that while the feature structures at the
root of $\tau_3$ and $\tau_4$ are not compatible with $\boxed{1}$ and
$\boxed{2}$, they do agree with respect to those parts that are fully
expanded at $\tau$'s root node. The ``context-freeness'' of \plpatr\
means that given the three trees shown in Figures~\ref{fig-reca}
and~\ref{fig-recb} the tree shown in Figure~\ref{fig-recc} will also be
generated by the grammar.

\begin{figure}
\begin{center}
\setlength{\unitlength}{0.012500in}%
\begingroup\makeatletter\ifx\SetFigFont\undefined
% extract first six characters in \fmtname
\def\x#1#2#3#4#5#6#7\relax{\def\x{#1#2#3#4#5#6}}%
\expandafter\x\fmtname xxxxxx\relax \def\y{splain}%
\ifx\x\y   % LaTeX or SliTeX?
\gdef\SetFigFont#1#2#3{%
  \ifnum #1<17\tiny\else \ifnum #1<20\small\else
  \ifnum #1<24\normalsize\else \ifnum #1<29\large\else
  \ifnum #1<34\Large\else \ifnum #1<41\LARGE\else
     \huge\fi\fi\fi\fi\fi\fi
  \csname #3\endcsname}%
\else
\gdef\SetFigFont#1#2#3{\begingroup
  \count@#1\relax \ifnum 25<\count@\count@25\fi
  \def\x{\endgroup\@setsize\SetFigFont{#2pt}}%
  \expandafter\x
    \csname \romannumeral\the\count@ pt\expandafter\endcsname
    \csname @\romannumeral\the\count@ pt\endcsname
  \csname #3\endcsname}%
\fi
\fi\endgroup
\begin{picture}(183,65)(168,600)
\thinlines
\put(200,640){\line(-2,-3){ 20}}
\put(180,610){\line( 1, 0){ 40}}
\put(220,610){\line(-2, 3){ 20}}
\put(320,640){\line(-2,-3){ 20}}
\put(300,610){\line( 1, 0){ 40}}
\put(340,610){\line(-2, 3){ 20}}
\put(220,600){\makebox(0,0)[b]{\smash{\SetFigFont{10}{12.0}{rm}$a_q$}}}
\put(180,600){\makebox(0,0)[b]{\smash{\SetFigFont{10}{12.0}{rm}$a_p$}}}
\put(200,620){\makebox(0,0)[b]{\smash{\SetFigFont{10}{12.0}{rm}$\tau_3$}}}
\put(200,655){\makebox(0,0)[b]{\smash{\SetFigFont{8}{9.6}{rm}$\ftermfsa$}}}
\put(340,600){\makebox(0,0)[b]{\smash{\SetFigFont{10}{12.0}{rm}$a_s$}}}
\put(300,600){\makebox(0,0)[b]{\smash{\SetFigFont{10}{12.0}{rm}$a_r$}}}
\put(320,620){\makebox(0,0)[b]{\smash{\SetFigFont{10}{12.0}{rm}$\tau_4$}}}
\put(320,655){\makebox(0,0)[b]{\smash{\SetFigFont{8}{9.6}{rm}$\gtermfsa$}}}
\end{picture}
\end{center}
\caption{Compatible subderivations}
\label{fig-recb}
\end{figure}
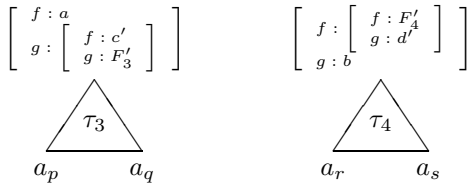

This gives us a means of efficiently storing the potentially unbounded
feature structures associated with nodes in a derivation tree (derived
feature structures). By analogy with the situation for \lig, derived
feature structures can be viewed as consisting of a bounded part
(relevant to rule application) plus unbounded information about the
values of features.  For each feature, we store in the recognition array
a bounded amount of information about its value locally, together with a
pointer to a further array element. Entries in this element of the
recognition array that are compatible (i.e. unifiable) with the bounded,
local information correspond to different possible values for the
feature. For example, we can use a single entry in the recognition array
to store the fact that all of the feature structures that can appear at
the root of the trees in Figure~\ref{fig-recc} derive the substring
$a_i\ldots a_j$. This entry would be underspecified, for example, the
value of feature $\boxed{1}$ would be specified to be any feature stored
in the array entry for the substring $a_p\ldots a_q$ whose feature $f$
had the value $a$.

However, this is not the end of the story. In contrast to \lig, \plpatr\
licenses structure sharing on the right hand side of productions. That
is, partial linearity permits feature values to be shared between
daughters where they are not also shared with the mother.  But in that
case, it appears that checking the applicability of a production at some
point in a derivation must entail the comparison of structures of
unbounded size.  In fact, this is not so.  The \plpatr\ recognition
algorithm employs a second array (called the compatibility array), which
encodes information about the compatibility of derived feature
structures.  Tuples of compatible derived feature structures are stored
in the compatibility array using exactly the same approach used to store
feature structures in the main recognition array.  The presence of a
tuple in the compatibility array (the indices of which encode which
input substrings are spanned) indicates the existence of derivations of
compatible feature structures.  Due to the ``context-freeness'' of
\plpatr, new entries can be added to the compatibility array in a
bottom-up manner based on existing entries without the need to
reconstruct complete feature structures.

\section{Conclusions}

In considering ways of extending \lig, this paper has introduced the
notion of partial linearity and shown how it can be manifested in the
form of a constrained unification-based grammar formalism.  We have
explored examples of the kinds of tree sets and string languages that
this system can generate. We have also briefly outlined the sense in
which partial linearity gives rise to ``context-freeness'' in derivations
and sketched how this can be exploited in order to obtain a tractable
recognition algorithm.

\section{Acknowledgements}

We thank Roger Evans, Gerald Gazdar, Aravind Joshi, Bernard Lang,
Fernando Pereira, Mark Steedman and K.~Vijay-Shanker for their help.

\end{document}